# A Novel CNN-LSTM-based Approach to Predict Urban Expansion

Wadii Boulila[1,2], Hamza Ghandorh[1], Mehshan Ahmed Khan[3], Fawad Ahmed[3], Jawad Ahmad[4]

[1]College of Computer Science and Engineering, Taibah University, Saudi Arabia
[2]RIADI Laboratory, University of Manouba, Tunisia
[3]Department of Electrical Engineering, HITEC University Taxila, Pakistan
[4]School of Computing, Edinburgh Napier University, United Kingdom

## Abstract

Time-series remote sensing data offer a rich source of information that can be used in a wide range of applications, from monitoring changes in land cover to surveilling crops, coastal changes, flood risk assessment, and urban sprawl. This paper addresses the challenge of using time-series satellite images to predict urban expansion. Building upon previous work, we propose a novel two-step approach based on semantic image segmentation in order to predict urban expansion. The first step aims to extract information about urban regions at different time scales and prepare them for use in the training step. The second step combines Convolutional Neural Networks (CNN) with Long Short Term Memory (LSTM) methods in order to learn temporal features and thus predict urban expansion. In this paper, experimental results are conducted using several multi-date satellite images representing the three largest cities in Saudi Arabia, namely: Riyadh, Jeddah, and Dammam. We empirically evaluated our proposed technique, and examined its results by comparing them with state-of-the-art approaches. Following this evaluation, we determined that our results reveal improved performance for the new-coupled CNN-LSTM approach, particularly in terms of assessments based on Mean Square Error, Root Mean Square Error, Peak Signal to Noise Ratio, Structural Similarity Index, and overall classification accuracy.

## 1 Introduction

The analysis of time-series data generally deals with high-temporal and low-spatial resolutions [1]. Economics [2], computer vision [3], remote sensing [4], and financial data [5] are just some of the many fields in which this analysis has been applied. Frequently, time-series data can provide better analysis than other datasets because it includes temporary changes at a particular period. In the field of remote sensing, satellites revisit the same place at different times and capture high-spatial-resolution images at consecutive times, thus generating time-series remote sensing (TSRS) [6]. These remote sensing data is a cost-effective mean of providing spatial information and temporal patterns regarding specific areas on the earth's surface. In particular, an analysis conducted via time-series satellite images at key dates can offer a surprisingly full and comprehensive characterization of land cover.

TSRS data have long been used as a tool for large-scale monitoring used in change prediction that focuses on vegetation, mapping of forest disturbances, analysis of forest degradation, land cover mapping, disaster forecasting and tracking, and many other areas [7-15]. Analyzing TSRS can offer vital insights across various sectors, from scientific progress to economic growth and into political developments. This phenomenon occurs because a timely, precise analysis of TSRS data affects the efficacy and accuracy of every application built upon such data. Nowadays, remote sensing (RS) data are mostly utilized by governments, decision-makers, and planners as a means of implementing future measures [16,17]. Prediction of Land Cover Changes (LCC) in many countries is based on on-the-ground observation; however, gathering data in this way is very costly and time-consuming. The most cost-effective and practical way of predicting land cover maps in both regional and individual fields is through the use of RS images. The analysis of RS data is performed to analyze mechanical disturbances such as changes in ecosystems, natural resources, living space, and biophysical environments of the earth's surface, which are usually caused by human activity [18].

Over the past few decades, satellites have been equipped with hundreds of sensors to capture tremendous amounts of TSRS data with high resolutions in spatial, temporal, and spectral dimensions. Many algorithms have been developed to analyze the TSRS data and extract meaningful information and insights from the data that describe changes and developments in land covers. The analysis of RS images is challenging due to its high-dimensionality and large volumes of these images. Changes in land cover can be mapped using image processing techniques, but traditional computer-vision algorithms only explore a tiny portion of the information provided in satellite data due to the high resolutions of these data [19].

Recently, due to progress in Artificial Intelligence (AI) prediction models developed from Machine Learning (ML), particularly Deep Learning (DL), much attention has been devoted to analyzing RS data. Among deep learning methods, certain models—such as Convolutional Neural Networks (CNN), Recurrent Neural Networks (RNN) [20], Long Short Term Memory (LSTM) [21], and Convolution LSTM (Conv-LSTM) [22]—are among those most commonly used [23-25]. Unlike traditional Computer Vision-based algorithms, these models rely on pre-defined structures and ready-labeled data, which can be used to learn independently; in this way, they do not require hand-tuning. Moreover, DL-based models can extract features automatically for a specific task (e.g., change prediction) by simultaneously training the model on a given dataset. These approaches capture the non-linear relationship between input and output by learning temporal and long-term dependencies through the use of multi-dimensional time-series data [26]. Given this structure, the accuracy of any given deep neural network depends on the network architecture it was based upon and the type of dataset used to train it. The performance of such a network can be enhanced by carefully selecting different processing units such as pooling, convolution, batch normalization, activation functions, and more, in the deep neural network layers [21].

This paper proposes a novel coupled CNN-LSTM approach to predict urban expansion in many regions of Saudi Arabia. The proposed approach is based on two steps:

- Semantic segmentation to extract urban regions from multi-date satellite images. This step also applies several operations to prepare the data gathered for use in training.
- Prediction of urban expansion using an approach coupling CNN and LSTM methods. This step aims to train the proposed ConvLSTM model to predict urban maps for upcoming dates.

The remainder of this paper is organized as follow. Related works are described in the literature review of Section 2, while the preliminaries section, which introduces unsupervised image segmentation, the ConvLSTM model, and the batch normalization is presented in Section 3. The dataset used to evaluate the performance of the proposed approach is emphasized in Section 4. Section 5 depicts the experimental results and the corresponding analysis, and Section 6 concludes the paper.

## 2 Literature review

RS data have been widely used in order to predict changes in LCC [27, 28], forest structure [29, 30], and vegetation [1, 31] at various scales and across an extensive variety of geographical locations. In order to predict changes in RS data, different forecasting methodologies based on statistical models [32, 33], hybrid models [34, 35], and Markov models [36, 37] have been developed by various researchers. Recently, AI-based prediction models have been used more often, given their ease of implementation and relative flexibility in handling huge amount of data.

Etemadi et al. [38] , for instance, analyzed LCC between soil, mangroves, and open water using Landsat satellite images. In this work, the CA-Markov model was applied to the Landsat satellite images to simulate and predict the LCC in mangrove forests for the year 2025. Experimental results demonstrated that 21ha area of mangrove growth has changed into open water by that date, while 28ha area was projected to expand in a landward direction.

Kumar et al. [38] studied the different explanatory variables associated with changes in forest cover (which included distances from roads, settlements, and the forest edge), all using the Logistic Regression Model (LRM) in tandem with Landsat satellite images. In their work, the LRM model offered successful predictions regarding forest and non-forest areas with a probability of 87%.

You et al. [40] introduced a method to predict crop yields using RS data. The proposed method uses a dimensionality reduction technique based on histograms to train the CNN and the LSTM. Gaussian Process layer was also incorporated on top of the neural networks model to overcome the limitation of spatio-temporal dependencies between data points. The proposed approach outperformed traditional methods by 30% with regard to Root Mean Squared Error (RMSE) and outperformed the same methods by 15% in terms of Mean Absolute Percentage Error (MAPE). Adnan et al. [41] calibrated the Snow Run-off Model (SRM) in tandem with MODIS satellite data as a means of predicting daily discharge from the Gilgit river. After calibrating the model's efficiency for the four years from 2007 through 2010, the proposed system reached a coefficient of model efficiency with 0.96, 0.86, 0.9, and 0.94 for each year, respectively.

Das et al. [42] presented the Deep-STEP approach, which they derived from the Deep Stacking Network (DSN) for spatiotemporal satellite RS data predictions. In this network, stacks of multilayer perception were used to model spatial features during a specific instance of time. With reference to RMSE as well as Mean Absolute Error (MAE), the proposed Deep-STEP approach produced significantly fewer errors as compared to more common approaches, including Multi-Layer Perceptron (MLP), DSN, and Nonlinear Autoregressive Neural Network (NARNET). Das et al. [43] later proposed a new variant of a deep recurrent neural network called the Self Adaptive Recurrent Deep Incremental Network Model (SARDINE), which was meant to nowcast from remotely sensed data. SARDINE can self-construct network architecture while also learning from data samples. Due to its self-constructing abilities, SARDINE saves significant computational time when dealing with large satellite images as compared to other models.

Pantazi et al. [44] used three supervised Self Organizing Maps (SOM)—namely, Counter Propagation Artificial Neural networks (CP-ANNs), XY-Fused Networks (XY-Fs), and finally, the Supervised Kohonen Networks (SKNs)— to handle multiple layers of data on the soil as well satellite imagery showing characteristics of crop growth, all in order to predict wheat yield within-field variations. Of these models, SKN achieved an overall accuracy rating of 81.65%, which was the best performance among the three.

Qi et al. [45] proposed a hybrid system based on the combination of Graph Convolutional networks with Long Short-Term Memory networks (GC-LSTM) as a means of predicting atmospheric spatiotemporal $PM_{2.5}$ concentration. In order to follow spatial dependencies among different stations, a graph convolution network was used to capture the temporal dependency of LSTM. Experimental results showed that the CG-LSTM performed better than RNN, backpropagation (BP) neural network, or convolution LSTM.

By investigating the literature, we note that few research works have been conducted to predict urban expansion. Majority of research has utilized the variants of Artificial Neural Networks (ANN), Markov Chain and statistical models to simulate and predict the Land Use and Land Cover (LULC) changes from RS data. However, most of existing research works have some limitations in terms of accuracy and the need of intervention of users in the training process. The emergence of DL and its ability to learn representations of data and spatial features automatically from multilayer network has provided great results in many fields such as image recognition, classification, segmentation, and prediction. The development of DL-based techniques to predict LULC changes based on time-series satellite images becomes possible. To the best of our knowledge, this study in the first work in the literature that propose a ConvLSTM model to understand and simulate the urban growth. Additionally, extraction of ground truth from the satellite images requires significant human, material resources, and the knowledge. In the proposed approach, we propose semi ground-truth extraction method using unsupervised image segmentation algorithm. Proposed ground extraction method automatically separates the different patterns from satellite image, which can then be easily separated to get the targeted area. These targeted are used as an input for the prediction algorithm.

## 3 Preliminaries

### 3.1 Unsupervised image segmentation

Image segmentation describes the process of splitting an initial image into non-overlapping regions so that each one can be localized in a meaningful way concerning its locations or intensities [46]. Most supervised segmentation algorithms require a significant quantity of labeled data to train before achieving the desired performances. For image segmentation, pixel-level annotation is a very tedious, labor-intensive, time-consuming, and expensive process. By contrast, unsupervised image segmentation algorithms are helpful in the absence of manually labeled data. Unsupervised image segmentation algorithms aim to partition the given image into some arbitrarily-given number of meaningful regions exclusive of any previous knowledge of labels [47]. In Computer Vision, a variety of algorithms are used for segmentation including K-Means [48], normalized cuts [49], graph cut [50], mean shift [51], and many more.

Kim et al. [52] recently used CNN and differentiable feature clustering for unsupervised image segmentation. In this model, no training data and labels are required to train the algorithm. Let $I = \{v_n \in R^3\}_{n=1}^{N}$ be the three-band satellite image where $N$ signifies the number of pixels found in a given input image. Each pixel in the image is normalized in the range [0,1]. Here, $\{x_n \in R^p\}$ is the p-dimensional feature vector, while $f$ and $g$ are the feature extraction and mapping function, respectively. Each pixel in the image is assigned with cluster labels $\{c_n \in Z\}$ by $c_n = g(x_n)$. Unknown labels $c_n$ are later predicted by training the parameters $f$ and $g$ have been trained without human supervision to guide that process. Figure 1 depicts the network architecture to be found in the unsupervised image segmentation algorithm used in our study.

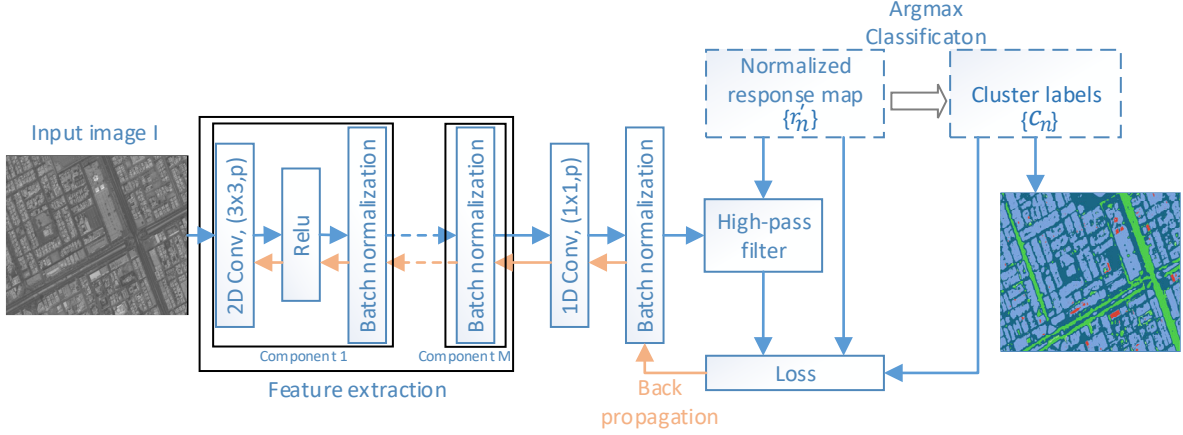

Figure 1: Unsupervised image segmentation algorithm network architecture [45].

Each convolutional component in the proposed architecture consists of three elements —a 2D convolution, a ReLU activation function, and finally a batch normalization function—to compute the p-dimensional feature map $\{x_n\}$ from $\{v_n\}$. In all M-convolutional components, $p$ filters of 3 x 3 pixels each are set to extract features. A linear classifier is used to obtain a response map $\{r_n = W_c x_n\}$ which is then normalized $\{r'_n\}$ such that $\{r'_n\}$ has zero mean and unit variance. Value cluster labels are acquired through the selection of a maximum value of $\{r'_n\}$ (argmax classification). Loss function $L$ of the network comprises two constraints; the similarity among features and the spatial continuity. The values of cluster labels and loss function are defined as follows [52]:

$$C_i = \{r'_n \in R^q | r'_{n,i} \geq r'_{n,j}, \forall_j\} \qquad 3.1$$
$$L = \underbrace{L_{sim}(\{r'_n, c_n\})}_{\text{Feature similarity}} + \underbrace{\mu L_{con}(\{r'_n\})}_{\text{Spatial continuity}} \qquad 3.2$$

In the above equations, $r'_{n,i} (i = 1, \ldots q)$ denotes the ith element of $r'_n$, where the quantity of distinct labels ranges from 1 to q. Weights for balancing of feature similarity and spatial continuity are denoted by μ. The loss function defined in 3.2 is calculated and backpropagated to update the convolutional filters $\{W_m\}_{m=1}^{M}$ and classifier $W_c$ parameters. During the training, network parameters are initialized with the Xavier initialization [53], and updated through stochastic gradient descent plus momentum.

## 3.2 Convolution LSTM

LSTM networks 23] were proposed to solve vanishing in RNNs [22]. Across the sequence of input data, LSTM networks can maintain a cell state $C_t$ from previous observations while also eliminating irrelevant information. To learn temporal features, LSTM networks take vectorized features maps as input, which are then encoded through fully connected layers. As a result of this vectorization, spatial correlation information is lost. On the other hand, CNNs are widely used to learn the relationship among the pixels of an input image by extracting feature maps using convolution operations.

Convolution LSTM (ConvLSTM) was proposed by Shi et al. [24] as an extension of Fully Connected (FC) LSTM [54] by replacing fully-connected operators with convolution operators to process sequential images. Convolutional operators in ConvLSTM add more computing power to the model with fewer parameters than conventional LSTM. ConvLSTMs can learn the global spatio-temporal information without shrinking the size of spatial feature maps. The process of ConvLSTMS can be formulated as follows [24]:

$$i_t = \sigma(W_{xi} * X_t + W_{hi} * H_{t-1} + W_{ci} \circ C_{t-1} + b_i), \quad 3.3$$
$$f_t = \sigma(W_{xf} * X_t + W_{hf} * H_{t-1} + W_{cf} \circ C_{t-1} + b_f), \quad 3.4$$
$$o_t = \sigma(W_{xo} * X_t + W_{ho} * H_{t-1} + W_{co} \circ C_{t-1} + b_o), \quad 3.5$$
$$\widetilde{C}_t = \tanh(W_{xc} * X_t + W_{hc} * H_{t-1} + b_c), \quad 3.6$$
$$C_t = f_t \circ C_{t-1} + i_t \circ \widetilde{C}_t \quad 3.7$$
$$H_t = o_t * \tanh(C_t) \quad 3.8$$

In the above equations, '$*$' denotes the convolution operator and '$\circ$' denotes the Hadamard product. Cell states are denoted by $C_1, \ldots, C_t$ and hidden states are denoted by $H_1, \ldots, H_t$, while the input gate is denoted by $i_t$, the forget gate is denoted by $f_t$, and the output is denoted by $o_t$. $\sigma$ is the sigmoid activation function that modulates gate output between the values of 0 and 1. $W_{xi}, W_{hi}, W_{ci}, W_{xf}, W_{hf}, W_{cf}, W_{xo}, W_{ho}, W_{co}, W_{xc}, W_{hc}$ are the 2D convolution kernels. $b_i, b_f, b_c$ and $b_o$ are the bias terms.

## 3.3 Batch Normalization (BN)

Batch normalization (BN) [55] describes a network re-parameterization method that assists in accelerating training by standardizing the distribution of inputs for each layer. BN preserves the activation of all layers with a mean of 0 and a variance of 1. The transformed outputs produced by BN are depicted as follows:

$$BN(X) = \gamma \frac{X - E[X]}{\sqrt{Var[X] + \epsilon}} + \beta \quad 3.9$$

Where X signifies the input to the batch normalization layer, $\epsilon$ signifies a minor constant included to avoid numerical instability, $E[X]$ and $Var[X]$ signify the mean and the variance, respectively, of the mini-batch. $\gamma$ and $\beta$ are the two learnable parameters.

# 4 Dataset

In order to establish and validate the effectiveness of our proposed approach, three regions in Saudi Arabia— namely, Riyadh, Jeddah, and Dammam—have been selected as a study site. These regions are among the largest cities in Saudi Arabia and constitute the country's most notable urban growth, making them an ideal case to gauge the effectives of the proposed scheme.

Riyadh is the capital city and central financial hub of Saudi Arabia. It is also considered one of the fastest-growing cities in the entire region of the Middle East. Meanwhile, Jeddah is a modern seaport and Saudi Arabia's second-largest city. Finally, Dammam is both the capital of Saudi Arabia's Eastern Province and also the center of the Saudi oil industry. According to [56, 57], about 40% of land cover for these three regions has been changed to urban areas between 2006 and 2013, making them a timely and relevant example for studying change in the region.

Our main observation is that Riyadh, Jeddah, and Dammam witnessed a rapid urban expansion in recent years. The main reason for the urban expansion is the significant socioeconomic growth that occurred in the country due to the rising of oil revenues.

In this section, experiments will be conducted using a three-part time-series of satellite imagery captured by the French *Satellite pour l'Observation de la Terre* (SPOT) satellite.

Each of the three time-series is composed of four images with a temporal frequency of two years. Table 1 depicts the region name, acquisition date, resolution, image vertex location, and sensor type of satellite images used to validate the proposed approach. These images have been corrected both radiometrically and geometrically using ortho-rectification and spatial registration with sub-pixel accuracy and comparing them against a global reference system. According to domain experts, there are six types of land cover to consider on these sites: these six are water, urban, soil, vegetation, mountain, and road.

In this paper, our CNN-LSTM method's performance is evaluated using a ground truth generated by domain experts as an assessment. An unsupervised image segmentation-based algorithm is used alongside Google Maps to generate the ground truth. We extracted only the areas that included clusters of buildings and houses from the satellite images.

A number of experiments have been conducted as a means of predicting urban change across these three selected regions.

*Table 1: Characteristics of satellite images.*

| Region Name | Acquisition date | Resolution | Image Vertex Location | Sensor type |
|---|---|---|---|---|
| Dammam | 2013-12-23 | 29893x31104 | (N026°49'22'' E049°39'36'')<br>(N026°40'44'' E050°17'23'')<br>(N026°09'65'' E050°08'27'')<br>(N026°17'42'' E049°30'47'') | SPOT5 |
| | 2015-05-25 | 20854x16790 | (N026°33'37" E050°00'34")<br>(N026°33'37" E050°14'34")<br>(N026°16'15" E050°14'34")<br>(N026°16'15" E050°00'34") | SPOT7 |
| | 2017-04-11 | 20051x16291 | (N026°33'16" E050°00'44")<br>(N026°33'16" E050°14'19")<br>(N026°16'33" E050°14'19")<br>(N026°16'33" E050°00'44") | SPOT6 |
| | 2019-02-04 | 20051x16291 | (N026°33'16" E050°00'44")<br>(N026°33'16" E050°14'19")<br>(N026°16'33" E050°14'19")<br>(N026°16'33" E050°00'44") | SPOT6 |
| Jeddah | 13 06 2013 | 29511x35521 | (N021°50'48" E039°00'14")<br>(N021°42'32" E039°44'56") | SPOT5 |

|  | 2015-04-22 | 29487x17003 | (N021°10'44" E039°36'56")<br>(N021°19'01" E038°53'26")<br>(N021°45'24" E039°03'28")<br>(N021°45'24" E039°17'38")<br>(N021°20'49" E039°17'38")<br>(N021°20'49" E039°03'28") | SPOT6 |
|---|---|---|---|---|
|  | 2017-07-08 | 29487x17003 | (N021°45'24" E039°03'28")<br>(N021°45'24" E039°17'38")<br>(N021°20'49" E039°17'38")<br>(N021°20'49" E039°03'28") | SPOT6 |
|  | 2019-01-11 | 28926x16646 | (N021°45'01" E039°03'39")<br>(N021°45'01" E039°17'31")<br>(N021°20'55" E039°17'31")<br>(N021°20'55" E039°03'39") | SPOT7 |
| Riyadh | 13/03/01 | 29593x31202 | (N025°02'49" E046°29'47")<br>(N024°53'03" E047°07'44")<br>(N024°22'24" E048°58'58")<br>(N024°31'08" E046°21'10") | SPOT5 |
|  | 2015-02-21 | 6570x7993 | (N024°44'45" E046°37'04")<br>(N024°44'45" E046°43'44")<br>(N024°39'17" E046°43'44")<br>(N024°39'17" E046°37'04") | SPOT6 |
|  | 2017-04-11 | 6619x8192 | (N024°44'46" E046°36'58")<br>(N024°44'46" E046°43'47")<br>(N024°39'15" E046°43'47")<br>(N024°39'15" E046°36'58") | SPOT6 |
|  | 2019-01-15 | 6419x8016 | (N024°44'40" E046°36'58")<br>(N024°44'40" E046°43'39")<br>(N024°39'19" E046°43'39")<br>(N024°39'19" E046°36'58") | SPOT7 |

# 5 Proposed approach

## 5.1 Image segmentation

The considered images are first segmented using the unsupervised image segmentation algorithm proposed by Kim et al. [52]. To apply the unsupervised segmentation algorithm, satellite images are divided into non-overlapping blocks of $1024 \times 1024$ pixels. A block-based approach is used for the segmentation in order to reduce the computation load using a single GPU. To perform this segmentation, each block is given as input to the algorithm proposed by Kim et al., as shown in Figure 1. The result of the segmentation of a single $1024 \times 1024$ image block is depicted below in Figure 2.

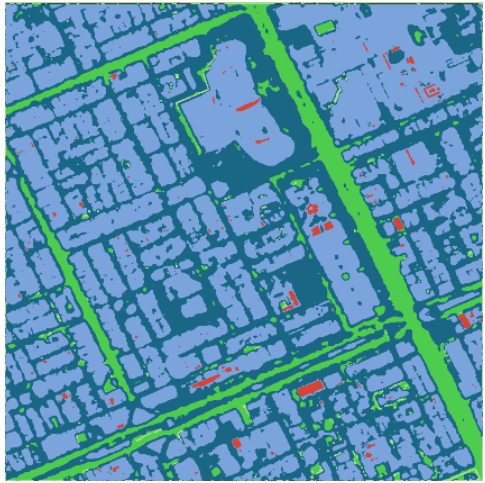

*Figure 2: Segmentation of a single $1024 \times 1024$ image block.*

The primary phases of the unsupervised image segmentation algorithm have been laid out in Section 3.1. As depicted in Figure 2, all six classes of land cover are depicted in different colors, but our target mask is only the blue, given that our purpose here is to track the progress of urban areas. Therefore, non-urban areas in each image are removed. Several operations, such as bwmorph and bwclean functions in Matlab, are then applied to the segmented images to obtain the final output results. After removing all the noise using such Matlab functions, we can note that the segmented images still contain some objects (highlighted in red in Figure 3) that are not part of buildings and houses. These unwanted objects are removed manually using the Region of Interest (ROI) tool in ENVI[1] to make useful data for training the neural network, as shown in Figure 4.

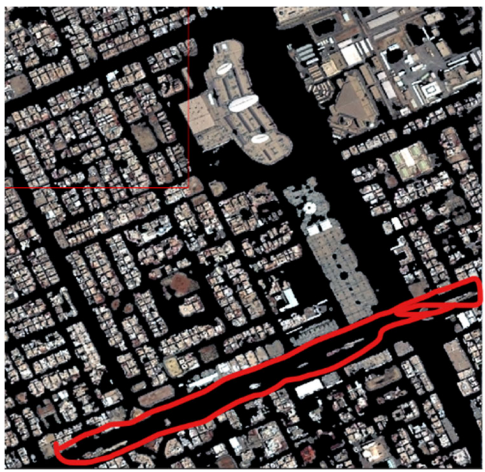

*Figure 3: Minor area remained after removing objects.*

---

[1] https://www.ittvis.com/envi/

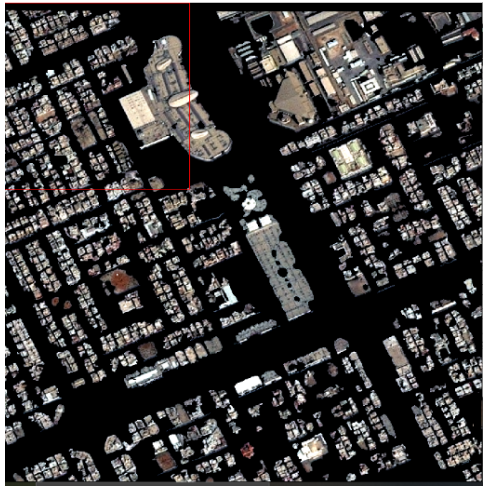

*Figure 4: All the noise removed using the ENVI tool.*

## 5.2 Prediction of urban Expansion

To train the proposed ConvLSTM model, a sequence of 3 satellites images $I_1$, $I_2$ and $I_3$ for the years 2015, 2017, and 2019 were used. The proposed architecture for the LCC prediction is shown in Figure 5. There are 4 ConvLSTM layers in this architecture, each with a kernel size of $3 \times 3$ and 40 filters, three batch normalization (BN) layers, and one convolution layer for the output.

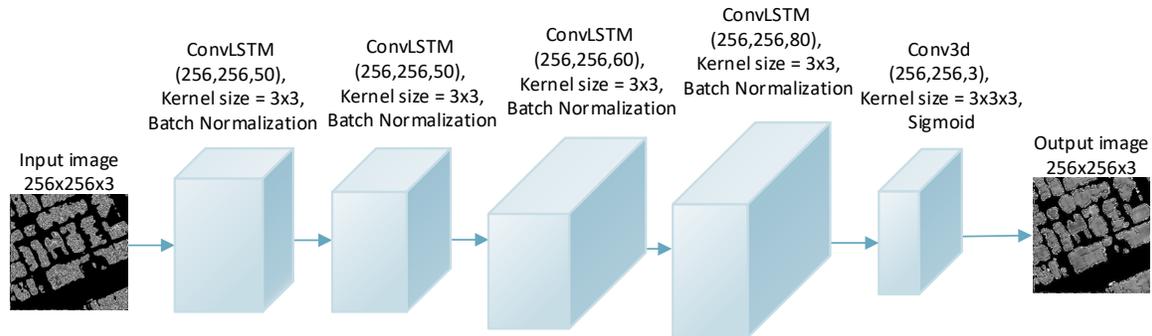

*Figure 5: The proposed architecture of ConvLSTM.*

The model is trained by randomly initializing the weights W. The model's input and output sizes are each $256 \times 256 \times 3$. An Adam optimizer [58] with default parameters (learning rate = 0.001, $\beta_1 = 0.9$, $\beta_2 = 0.999$) was applied. The weights of the model were updated by minimizing categorical cross-entropy loss. The values of hyper-parameters in the ConvLSTM are listed in Table 2.

*Table 2: Hyperparameters in ConvLSTM*

| Hyperparameters | Value |
|---|---|
| Batch size | 10 |
| Learning rate | 0.001 |
| Optimizer | Adam [51] |
| $\beta_1$ | 0.9 |
| $\beta_2$ | 0.999 |

| Input image size | 256 × 256 × 3 |
|---|---|
| Loss function | Categorical cross-entropy |
| Feature scaling | [0,1] |
| Hidden layers | 5 |
| Activation functions | tanh (ConvLSTM) sigmoid (Conv3d) |

The RS images used in this paper cannot be processed directly due to the high spatial resolution. Therefore, the RS images are divided into n non-overlapping blocks of size 256 × 256 pixels.

Let $B_1$, $B_2$ and $B_3$ be the vectors containing the blocks of satellite images $I_1$, $I_2$ and $I_3$, respectively.

$$B_1 = B_1^1, B_2^1, B_3^1 \ldots, B_n^1 \qquad 5.1$$
$$B_2 = B_1^2, B_2^2, B_3^2 \ldots, B_n^2 \qquad 5.2$$
$$B_3 = B_1^3, B_2^3, B_3^3 \ldots, B_n^3 \qquad 5.3$$

Each block of the vector contains the normalized pixel values $a_{1,1}, a_{1,2}, \ldots, a_{128,128}$ of the satellite image in the range between 0 to 1. Mathematically, each block is denoted as follows:

$$B_j^i = \begin{bmatrix} a_{1,1} & \cdots & a_{1,256} \\ \vdots & \ddots & \vdots \\ a_{256,1} & \cdots & a_{256,256} \end{bmatrix} \text{ where } 1 \leq i \leq 3 \text{ and } 1 \leq j \leq n \qquad 5.4$$

For training blocks of images, the satellite images of the year 2015 are used as input X of our ConvLSTM model, and blocks of 2017 images were used as the output to the model Y, as demonstrated by the following equation.

$$X = [X_1, X_2, X_3, \ldots, X_n] \qquad 5.5$$
$$Xj = B_j^k \qquad 5.6$$
$$Y = [B_1^m, B_2^m, B_3^m \ldots, B_n^m] \qquad 5.7$$

For training k=1 and m=2 are selected in the above equations, while for the validation, $k = 2$ and $m = 3$ are selected while $1 \leq j \leq n$. The pseudo-code used by our proposed approach is outlined below in Algorithm 1.

---
**Algorithm 1.** Training of the ConvLSTM Model.

**Input:**
   $I_1$, $I_2$ and $I_3$ : Satellite images for the years 2015, 2017, and 2019.
**Output**:
   ***Trained Model:*** ConvLSTM-based model (to be used for predicting the next sequence of the block).
1. Divide the images $I_1$, $I_2$, and $I_3$ into **256 × 256** blocks.
   $B_1 = B_1^1, B_2^1, B_3^1 \ldots, B_n^1$
   $B_2 = B_1^2, B_2^2, B_3^2 \ldots, B_n^2$
   $B_3 = B_1^3, B_2^3, B_3^3 \ldots, B_n^3$
2. Prepare the input **X** and the output **Y** for training and validation.
   $X = [X_1, X_2, X_3, \ldots, X_n]$

$$Xj = B_j^k$$
$$Y = [B_1^m, B_2^m, B_3^m \ldots, B_n^m]$$
**For training:** $k = 1$ and $m = 2$
**For validation:** $k = 2$ and $m = 3$
3. Randomly initialize weights $W$ and select the batch size 10 to train the ConvLSTM-based model.
4. Train the model by updating weights $W$ and other model parameters by minimizing categorical cross-entropy loss using Adam optimizer.

# 6 Experimental results

As mentioned earlier, we split up each satellite image dataset into non-overlapping blocks measuring $256 \times 256$ pixels. For the training, all the image blocks of the year 2015 were used as input, while the image blocks of the year 2017 wre used as the output. For validation, all image blocks of the year 2017 were given to the model, and the resultant images were compared with the image blocks of the year 2019, as shown in Figure 6. In this study, experimental results are conducted using a machine with four core CPU, 16GB RAM, and Nvidia Tesla T4 GPU.

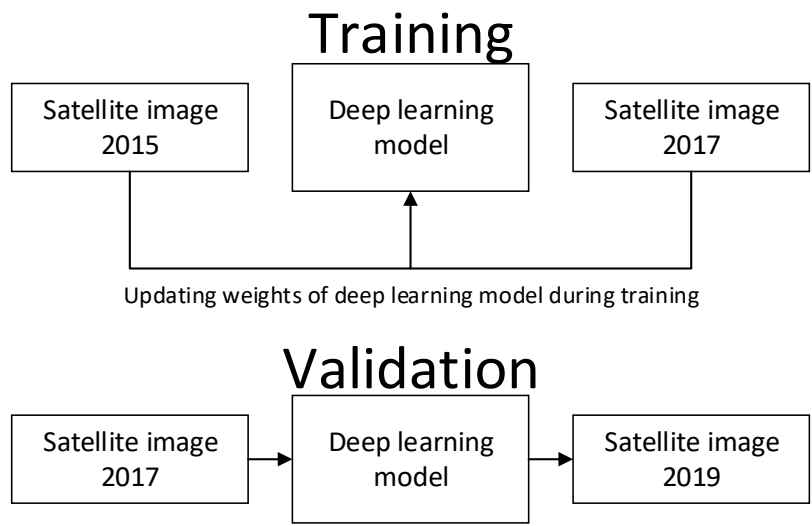

Figure 6: ConvLSTM model training and validation.

We trained the ConvLSTM model on our dataset before assessing its performance. The ConvLSTM model is trained with a categorical cross-entropy loss, batch size of 10, and the Adam optimizer. The proposed model is trained up to 32 epochs (i.e., after reaching the 32$^{nd}$ epoch, the accuracy is not further increased and the loss is not further decreased). The results of training and validation are shown in Figure 7. We notice that proposed ConvLSTM based approach gives maximum accuracy of 92.19% with a categorical cross-entropy loss of 0.242 on the validation data.

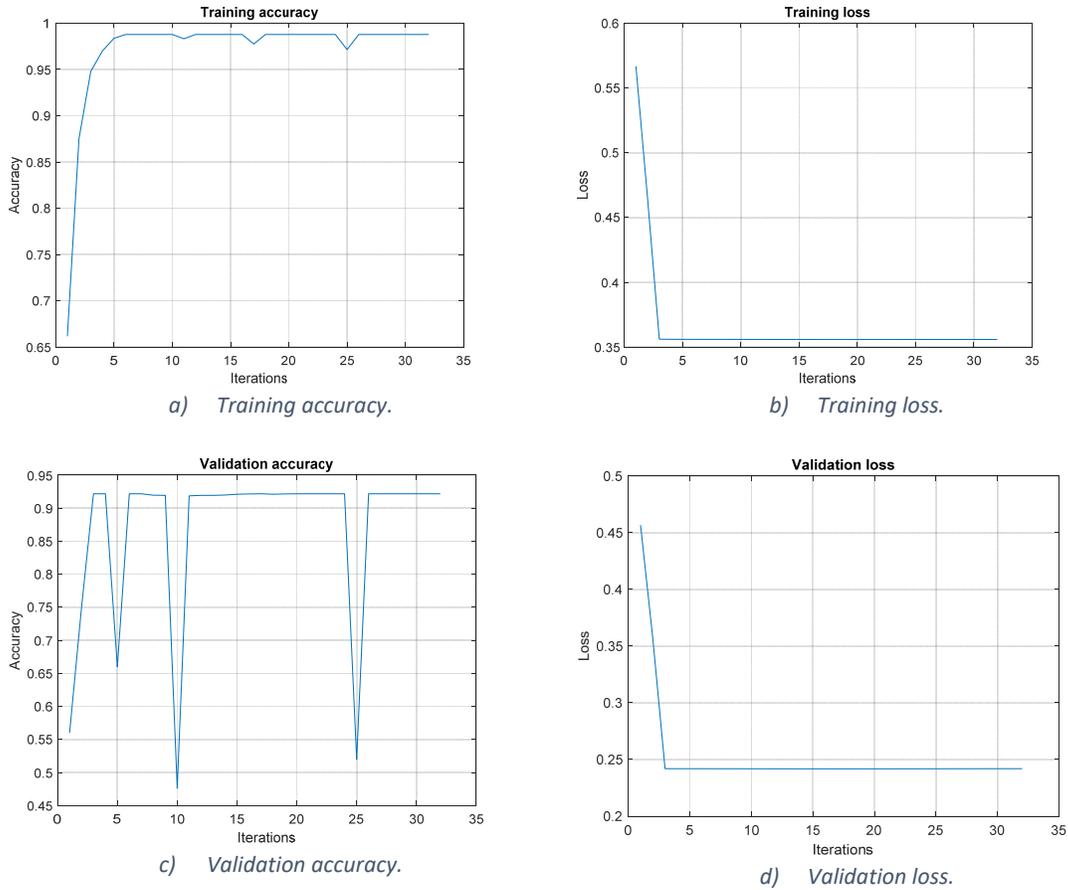

*Figure 7: Accuracy and loss plot of ConvLSTM*

We then trained the Pix2Pix GAN [59] and Dual GAN [60] in a similar manner. The Pix2pix model is trained for 35 epochs, and the Dual GAN model is trained for 37 epochs until each reached the minimum loss threshold. The training time of each algorithm is shown in Table 3. We notice that Pix2Pix GAN provides the best training time with 3.16 hours, followed by the proposed model with 4.17 hours and Dual GAN with 6.39 hours. The discriminator L1, L2 loss, and generator loss [59] of the Pix2pix GAN are shown in Figure 8 and the reconstruction losses [60] of dual GAN are shown Figure 9.

*Table 3: Training time.*

| Algorithm | Epochs | Training time |
|---|---|---|
| Pix2Pix GAN | 35 | 3.16hrs |
| Dual GAN | 37 | 6.39hrs |
| ConvLSTM (proposed) | 32 | 4.17hrs |

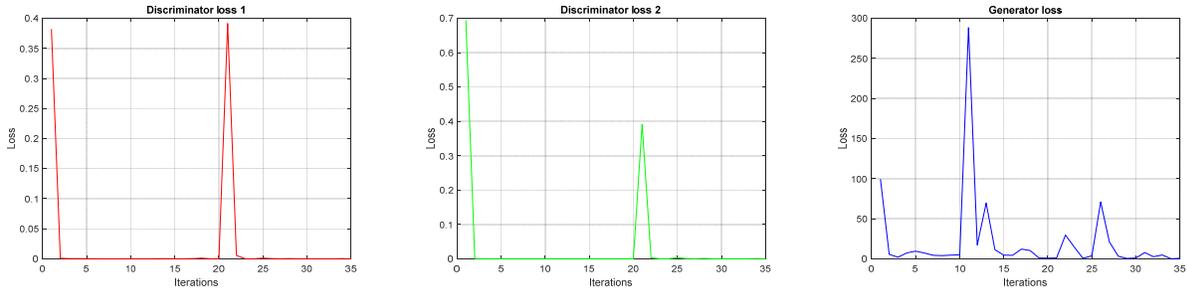

*Figure 8: Pix2pix GAN Losses.*

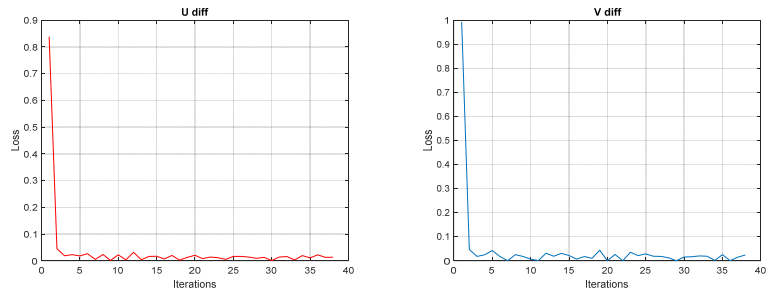

*Figure 9: Dual GAN Losses.*

In order to assess the quality of predictions offered by our proposed ConvLSTM model, evaluations are carried out using industry-standard metrics: Mean Square Error (MSE), Root Mean Square Error (RMSE), Peak Signal to Noise Ratio (PSNR), and the Structural Similarity Index (SSIM). Error values, PSNR and SSIM of Pix2pix and Dual GAN are almost similar in Figure 10, Figure 11, Figure 12 and Figure 13. Lower values of MSE and RMSE indicate a better quality of prediction. The MSE and RMSE values obtained are depicted in Figure 10 and Figure 11, respectively. Although some values of Pix2pix and Dual GAN are equal to ConvLSTM but most error values from the ConvLSTM model are less compared to the errors of Pix2pix and Dual GAN.

The PSNR and SSIM values of the same image blocks are depicted in Figure 12 and Figure 13. Higher values of PSNR and SSIM mean better prediction results. Although some values of SSIM and PSNR in Pix2pix and Dual GAN are equal to ConvLSTM but most of the values obtained by ConvLSTM model are higher than those obtained by the Pix2pix and Dual GAN.

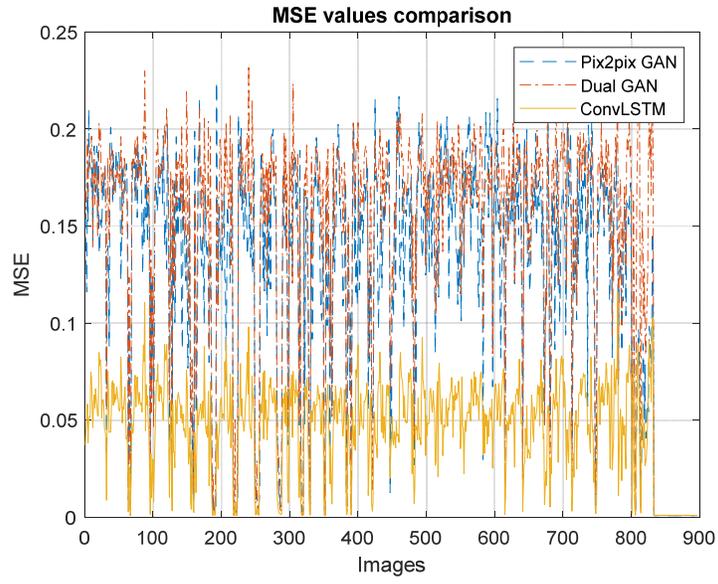

Figure 10: MSE values of validation image blocks.

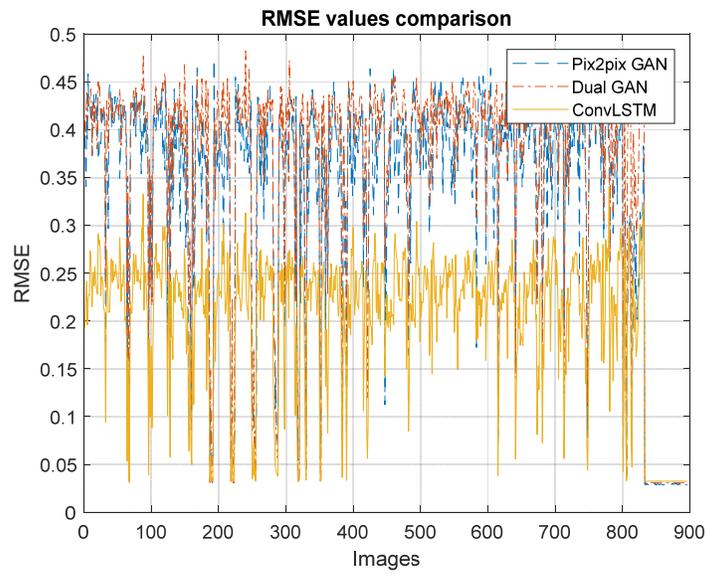

Figure 11: RMSE values of validation image blocks.

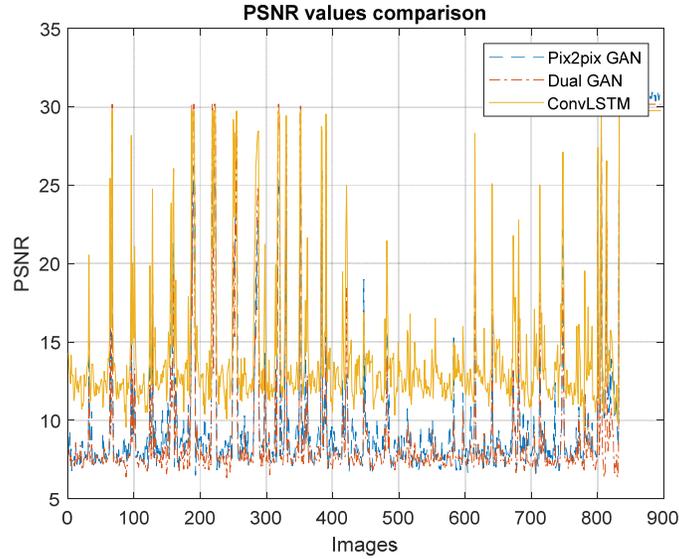

*Figure 12: PSNR values of validation image blocks.*

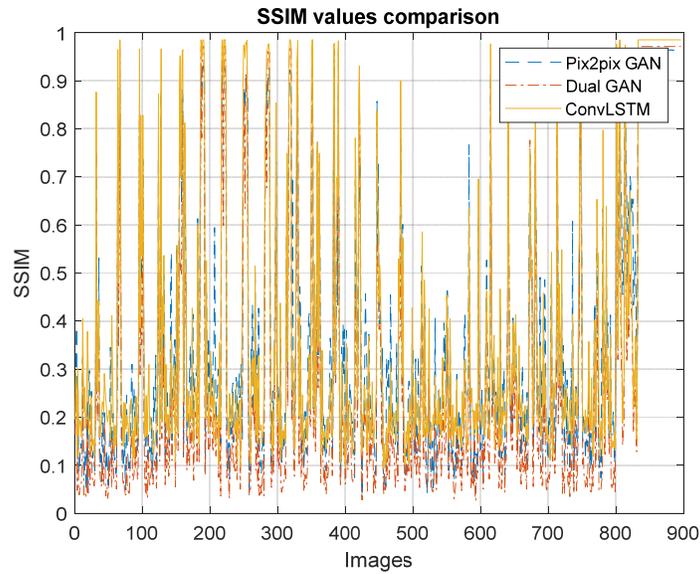

*Figure 13: SSIM values of validation image blocks.*

The mean values of MSE, RMSE, PSNR, and SSIM shown in Table 4 depict that the ConvLSTM model produces better results across all four indicators, as compared to the results obtained by the Pix2pix and Dual GAN models.

*Table 4: Metrics values comparison of ConvLSTM, Pix2pix GAN, and Dual GAN.*

|  | SSIM | PSNR | RMSE | MSE |
|---|---|---|---|---|
| **Pix2pix GAN** | 0.3601 | 10.9763 | 0.3342 | 0.1266 |
| **Dual GAN** | 0.2815 | 10.4498 | 0.3575 | 0.1437 |
| **ConvLSTM (proposed)** | **0.3694** | **15.1079** | **0.2021** | **0.0464** |

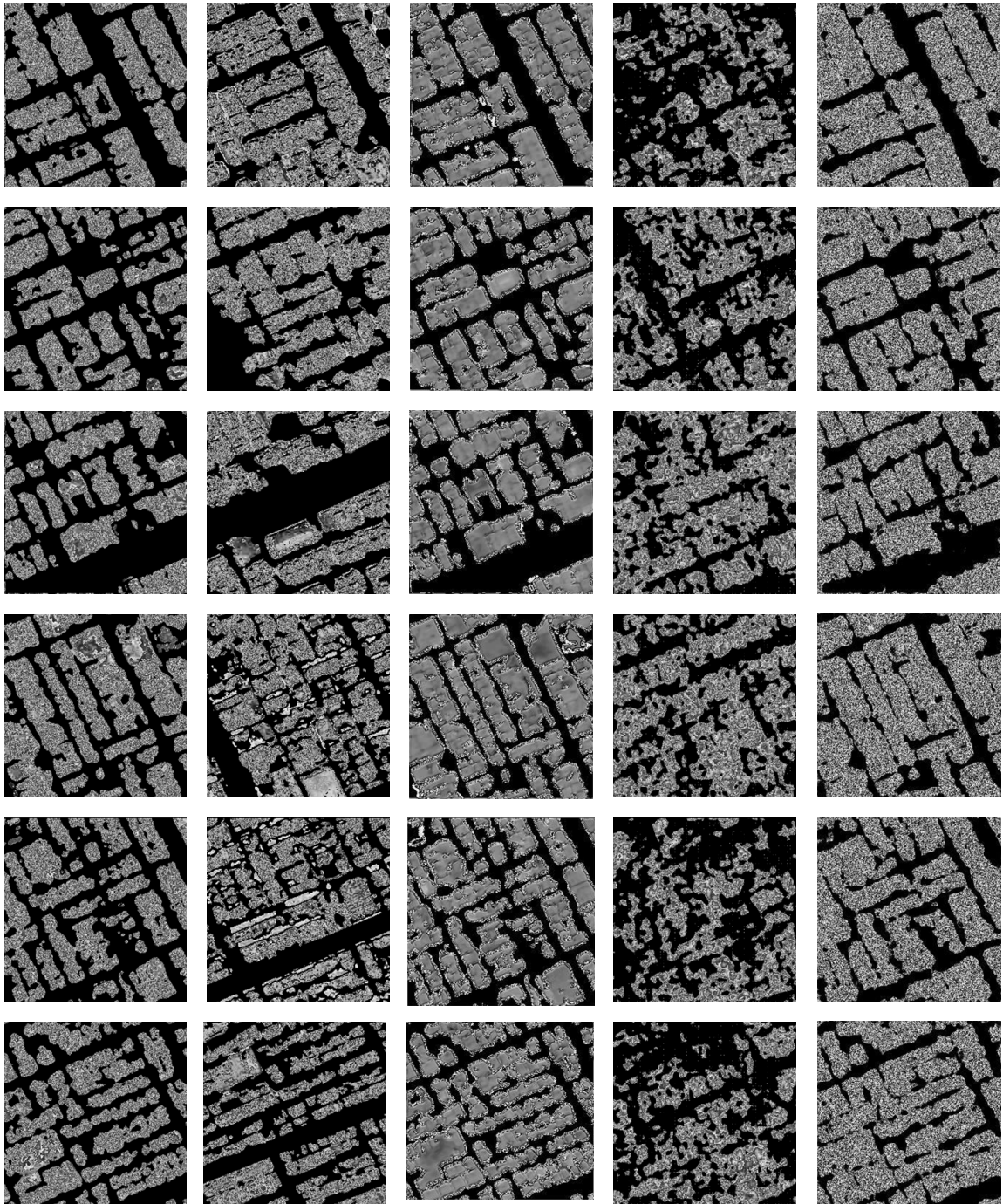

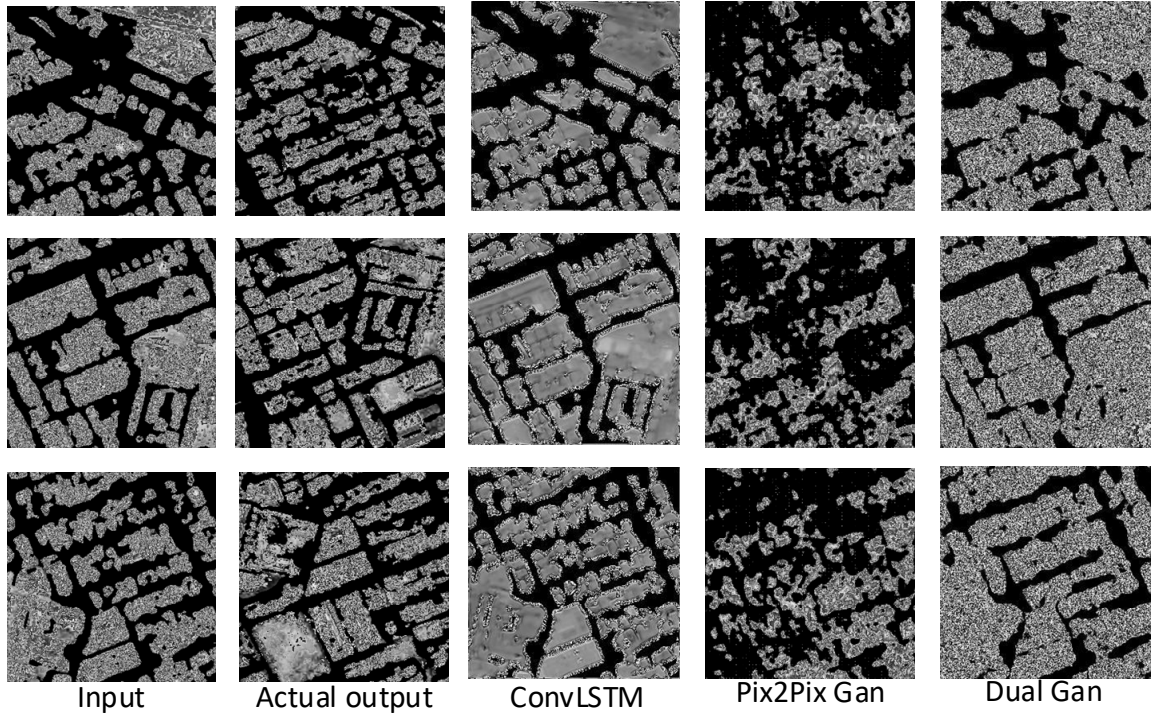

*Figure 14: Prediction results of ConvLSTM, Pix2pix GAN, and Dual GAN.*

Figure 14 shows another example of the application of the three prediction models used in this paper. The input images to the three prediction models are urban areas taken from the satellite image in the year 2017. The output of the models is urban areas representing the same regions in the year 2019. Actual output is the ground truth of the urban areas obtained from the year 2019.

According to the considered criteria (i.e., SSIM, PSNR, RMSE, and MSE), the proposed ConvLSTM model outperforms the two models. From a visual perspective, we can note that the Pix2Pix GAN model provides good results for non-urban areas. On the other hand, the Dual GAN outperforms Pix2Pix GAN in predicting urban areas but not better than the proposed ConvLSTM model.

In order to assess the results of prediction obtained by ConvLSTM, Pix2pix GAN, and Dual GAN models for the satellite image acquired on January 11, 2019, we use the confusion matrix as depicted in Figure 15. Rows represent the two classes (urban and non-urban) for the ground truth images and columns represent the two classes (urban and non-urban) for the predicted changes obtained by the three models. As illustrated in Table 5, RS-DCNN provides good image classification with an overall classification accuracy of 92.19% and a value of Kappa equal to 0.844. Pix2Pix GAN has an overall classification accuracy of 90.30% and Dual GAN has an overall classification accuracy of 88.88%.

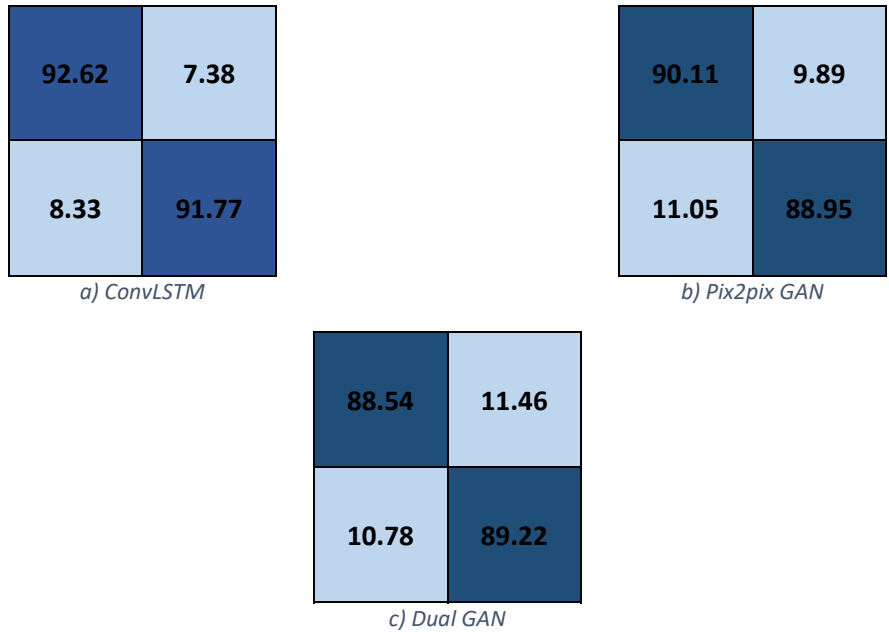

Figure 15: Normalized confusion matrix for ConvLSTM, Pix2pix GAN and Dual GAN models.

# 7 Discussions

In this study, we have proposed a more effective approach for predicting urban expansion in the Kingdom of Saudi Arabia (KSA). The proposed approach has been based upon fusing CNN and the LSTM models.

In order to train these models, semantic segmentation was conducted on a real dataset representing the three largest cities in Saudi Arabia. The dataset is comprised of several multi-date satellite images. In proposing, testing, and evaluating a new approach such as this one, we intend to provide insights, recommendations, and decision support for plausible future LCC in general and urban expansion in particular. The proposed solution will help to identify future probable risk areas at spatio-temporal scale.

The study of urban expansion will help authorities, regional planning, and decision-makers to create plans and policies that could help manage the urban expansion process.  Assessing urban growth can help in several environmental-related applications, such as developing effective strategies for urban planning, policy, and improved services for urbanization. Indeed, urban growth is considered as one of the important indicators of a country's economic development. Assessing urban expansion can help categorize urban sprawl, leading to converting natural and agricultural lands into urban. Urbanization can create serious environmental and socioeconomic challenges, including the loss of agricultural lands, the imbalance between the size of the urban regions and the needs of its population, the restriction of urban services such as water, roads, and sewers. Studying urban expansion also plays an essential role in preventing social problems such as declining social interaction, inadequate public services, deficient population health, and rising crime or delinquency. Likewise, understanding urban growth can help preserve life quality through factors such as air pollution, water quality, climate change, and public open spaces.

Saudi Arabia has undergone significant urban growth over the past five decades, moving rapidly from a predominantly rural society toward a majority urban one. This growth has been driven mainly by rapid economic growth from oil revenue. As a result, though, Saudi Arabia now ranks among the world's most urbanized countries, with more than 80% of its population living in urban areas [61]. This study has taken such realities into account, focusing on the three cities of Riyadh, Jeddah, and Dammam, each an urban center.

Riyadh has witnessed a circular urban expansion, starting from the center and expanding toward the north, west, and east. Meanwhile, Jeddah has witnessed urban expansion to the north and east more than the south. For Dammam, urban expansion has occurred west directly along with the Arabian Gulf. As we have noted here, in most cases, the urban expansion has occurred through existing urban land cover, rather than separate from it.

This continuous growth is far beyond the needs of both present and projected population sizes. Riyadh's population has grown from around 5.6 million in 2013 to approximately 7 million in 2019, while its urban areas have expanded more than three times in size. Meanwhile, Jeddah's population has grown from around 3.9 million in 2013 to approximately 4.5 million in 2019, and its urban areas have grown over twice in size even as its population growth was slower. Finally, Dammam's population has grown from approximately 0.7 million in 2013 to around 1.2 million in 2019. Here, the urban areas expanded by 400%, while its population did not even double.

A key challenge here is that Saudi Arabia's government must provide sufficient infrastructure—including basic services, housing, and jobs—for about 50% of its current population in these three areas alone. The results depicted in this study are in accordance with those obtained through previous studies [62,63].

When coupled with ineffective urban planning problems, the rapid urbanization in these three regions—Riyadh, Jeddah, and Dammam—has created many challenges for regional planning and decision-makers. In this regard, we propose that public awareness campaigns could effectively reach the appropriate entities. Saving urban space and combating urban sprawl will be crucial, so a focus on compact, high-density residential developments will be necessary to provide adequate housing and minimize transportation energy consumption. Overall, an approach to urban growth that focuses on high-density, mixed-land-use with high-rise architectures instead of villa-type housing is certainly required. Regional planning should also consider redesigning the existing building bylaws to accommodate more people in the same area.

## 8  Conclusion

In recent years, LCC has experienced important changes due to economic development and human activity. Monitoring LCC has become a hot research topic because having data in this area can help strategic land planning managers understand land use and land cover changes across vast areas. In particular, monitoring LCC help managers when considering major events such as landslides, erosion, land planning, and global change. In this paper, we proposed a new approach to study the urban expansion of the three largest cities in Saudi Arabia. Our proposed approach is based on two main steps: semantic image segmentation and prediction of urban expansion. When assessed against commonly used algorithms, our proposed method performed well in terms of MSE, RMSE, PSNR, SSIM, and overall classification accuracy. Based on the obtained results, we can assert that the proposed solution can be considered as a useful tool for sustainable urban planning.

An interesting perspective of the proposed work would be investigating the case of big remote sensing data. Indeed, the increasing volume of satellite images makes it very difficult to apply the proposed approach using a traditional computing framework. Recent research has been proposed to classify big remote sensing satellite images through a distributed CNN. From here, future research could also consider the case of securing big satellite images shared across several servers.

## Acknowledgments

The authors would like to thank King Abdul-Aziz City for Science and Technology (KACST) in Riyadh, Saudi Arabia for providing satellite data used in this study. In addition, the authors express their gratitude to the Editor-in-Chief, associate editors, and reviewers for their insightful comments and suggestions on the current work.

## Conflicts of Interest

The authors declare no conflicts of interest.

## Availability of data and material

Data will be available upon request to the corresponding author.